\documentclass[conference]{IEEEtran}
\usepackage{etoolbox}
\patchcmd{\thebibliography}{\section*{\refname}}{}{}{}

\usepackage{cite}
\usepackage{amsmath,amssymb,amsfonts}
\usepackage{graphicx}
\graphicspath{ {./figures/} }

\usepackage{textcomp}  %
\usepackage{nicefrac}
\usepackage{pifont}
\usepackage{soul}
\usepackage[font=footnotesize,labelfont=bf]{caption}
\usepackage{enumitem}
\usepackage[caption=false, font=footnotesize]{subfig}
\usepackage[table,xcdraw, dvipsnames]{xcolor}
\usepackage{titlesec}
\usepackage[normalem]{ulem}
\usepackage{wrapfig}
\usepackage{multirow}
\usepackage[export]{adjustbox}
\usepackage{stackengine}    
\usepackage{float}    
\usepackage[edges]{forest}
\usepackage{array}
\usepackage{stfloats}
\usepackage{tikz}
\usepackage{pgfplots}
\usepackage{caption}
\usepackage[outline]{contour}
\usepackage{microtype}
\usepackage{ragged2e}

\usepackage{tablefootnote}
\usepackage{tabularx}
\usetikzlibrary{tikzmark,patterns,shapes.misc,shapes.geometric,positioning,fit}

\usepackage[colorlinks = true,
            linkcolor = blue,
            urlcolor  = blue,
            citecolor = blue,
            anchorcolor = blue,
            pagebackref=true]{hyperref}

\definecolor{folderbg}{RGB}{124,166,198}
\definecolor{folderborder}{RGB}{110,144,169}
\definecolor{IGNGREEN}{RGB}{153, 211, 142}
\definecolor{TITLES}{RGB}{153, 211, 142}
\definecolor{TITLES_PRE}{RGB}{247, 212, 188}

\newlength\Size
\newenvironment{simplechar}{
   \catcode`\$=12
   \catcode`\&=12
   \catcode`\#=12
   \catcode`\^=12
   \catcode`\_=12
   \catcode`\~=12
   \catcode`\%=12
}{}

\def\BibTeX{{\rm B\kern-.05em{\sc i\kern-.025em b}\kern-.08em
    T\kern-.1667em\lower.7ex\hbox{E}\kern-.125emX}}
    
\tikzset{%
  folder/.pic={%
    \filldraw [draw=folderborder, top color=folderbg!50, bottom color=folderbg] (-1.05*\Size,0.2\Size+5pt) rectangle ++(.75*\Size,-0.2\Size-5pt);
    \filldraw [draw=folderborder, top color=folderbg!50, bottom color=folderbg] (-1.15*\Size,-\Size) rectangle (1.15*\Size,\Size);},
  file/.pic={%
    \filldraw [draw=folderborder, top color=folderbg!5, bottom color=folderbg!10] (-\Size,.4*\Size+5pt) coordinate (a) |- (\Size,-1.2*\Size) coordinate (b) -- ++(0,1.6*\Size) coordinate (c) -- ++(-5pt,5pt) coordinate (d) -- cycle (d) |- (c) ;},
}

\forestset{%
  declare autowrapped toks={pic me}{},
  pic dir tree/.style={%
    for tree={folder,font=\ttfamily,grow'=0,},
    before typesetting nodes={%
      for tree={edge label+/.option={pic me},},},
  },
  pic me set/.code n args=2{%
    \forestset{%
      #1/.style={inner xsep=2\Size,pic me={pic {#2}},}
    }
  },
  pic me set={directory}{folder},
  pic me set={file}{file},
}

\newenvironment{Tabular}[2][1]
  {\def\arraystretch{#1}\tabular{#2}}
  {\endtabular}

\renewcommand\thesubsection{\thesection.\arabic{subsection}}

\renewcommand\thesubsectiondis{\thesectiondis.\arabic{subsection}}

\begin{document}
\title{PureForest2024}
\author{Charles Gaydon}
\date{April 2024}
\twocolumn[{
  \begin{@twocolumnfalse}
    \vspace{-0.38cm}
    \begin{figure}[H]
    \centering
    \includegraphics[width=2.052\linewidth]{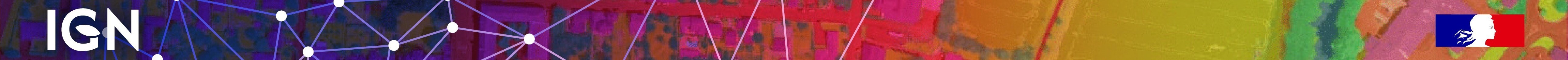}
    \end{figure}
    \parbox{\textwidth}{
      \centering
      \LARGE PureForest: A Large-Scale Aerial Lidar and Aerial Imagery Dataset for Tree Species Classification in Monospecific Forests 
    }\vspace{+0.4cm}
    \centering
    \large{Charles Gaydon \hspace{3cm} Floryne Roche}    \\
    \vspace{+0.2cm}
    \large{Institut national de l’information géographique et forestière (IGN), France} \\
    \vspace{+0.2cm}
    \normalsize{pureforest.dataset@ign.fr}    \\
    \vspace{+0.4cm}
  \end{@twocolumnfalse}
}]

\justifying
\setlength{\parskip}{2pt plus2pt}

\hspace{-1em}\textbf{Abstract}\textemdash\textit{Knowledge of tree species distribution is fundamental to managing forests. New deep learning approaches promise significant accuracy gains for forest mapping, and are becoming a critical tool for mapping multiple tree species at scale. To advance the field, deep learning researchers need large benchmark datasets with high-quality annotations.
To this end, we present the PureForest dataset: a large-scale, open, multimodal dataset designed for tree species classification from both Aerial Lidar Scanning (ALS) point clouds and Very High Resolution (VHR) aerial images. Most current public Lidar datasets for tree species classification have low diversity as they only span a small area of a few dozen annotated hectares at most. In contrast, PureForest has 18 tree species grouped into 13 semantic classes, and spans 339 km² across 449 distinct monospecific forests, and is to date the largest and most comprehensive Lidar dataset for the identification of tree species. By making PureForest publicly available, we hope to provide a challenging benchmark dataset to support the development of deep learning approaches for tree species identification from Lidar and/or aerial imagery. In this data paper, we describe the annotation workflow, the dataset, the recommended evaluation methodology, and establish a baseline performance from both 3D and 2D modalities.}

\begin{table}[h!]
\centering
\renewcommand{\arraystretch}{1.5}
\begin{Tabular}[1.5]{|p{8.2cm}|}
\hline \rowcolor[HTML]{a5b8eb} \textbf{Key Figures}                    \\  \hline
\end{Tabular}
\par\vskip1.2pt
\begin{Tabular}[0.8]{|p{8.2cm}|}
\hline

\rowcolor[HTML]{f7f9fd} \color{black}\ding{212} \color{black} 135,569 patches (50 m $\times$ 50 m), totalling 339 km² \\
\rowcolor[HTML]{f7f9fd} \color{black}\ding{212} \color{black} 449 distinct closed forests in 40 southern French departments \\
\rowcolor[HTML]{f7f9fd} \color{black}\ding{212} \color{black} 18 tree species grouped into 13 semantic classes \\
\rowcolor[HTML]{f7f9fd} \color{black}\ding{212}\color{black}Modality 1: colorized ALS point clouds (10 pulses/m², 40 pts/m²) \\
\rowcolor[HTML]{f7f9fd} \color{black}\ding{212}\color{black}Modality 2: VHR aerial images (250$\times$250 pixels at 0.2 m spatial resolution) \\[1em] \hline
\end{Tabular}
\end{table}

\vspace{-0.4cm}
\section{\justifying \textbf{Introduction}}
In Europe, forests are increasingly vulnerable and already subject to the consequences of climate change. Those include devastating fires, new parasites, physiological stress leading to higher mortality and reduced growth \cite{ForestEurope2020}. Monitoring tree species at a national scale is required to support public policies of forest management, for instance to promote more resilient tree species \cite{Keenan2015}.

Large-scale forest mappings typically involve visual identification for individual species determination. Species identification is complicated by the dependency of tree appearances to many factors such as stand age, stand structure, stand management, hydrologic and soil conditions, access to light, and season. Additionally, the aerial imagery data typically used for photointerpretation are subject to spatial or spectral perturbations, depending on the conditions of acquisition and post-processing: occlusions, weather, camera angles or radiometric correction.

Forest mapping therefore requires specialized knowledge and is time and labor-intensive. For instance, the French Mapping Agency (Institut national de l’information géographique et forestière, IGN) took more than a decade (from 2007 to 2018) to map the 171,000 km² of forest in metropolitan France \cite{bdforet}. To make accurate forests maps and keep them up-to-date, public agencies need to develop large-scale, automated methodologies.  

Research on tree species mapping is active. Recent work has focused on machine learning or deep learning from multispectral satellite imagery \cite{Holzwarth2020}. Very High Resolution (VHR) aerial imagery is less commonly used but has been successfully applied to species classification \cite{Holzwarth2020}. In particular, on a large reference dataset, the classification performance from aerial imagery outperformed mono-date satellite-based approaches by a large margin \cite{treesatai}.

Comparatively, deep learning approaches for large scale tree species mapping from ALS is an underexplored research area. ALS point clouds enable the extraction of features related to tree height, canopy density, and crown shape, which are discriminative features of tree species. And unlike aerial imagery, ALS is not affected by lighting conditions. 

Some small-scale studies showed that deep learning models can classify a limited number of species from Lidar data with reasonable accuracy. But these results lack robustness and generality due to the use of private datasets of insufficient scope (see Section \ref{related_works}). To overcome these limitations, researchers would greatly benefit from large ALS benchmark datasets with curated labels, semantic and spatial diversity, that are representative of a large area. Following recommended practices, benchmark datasets should be open with unrestricted access and have sufficient metadata including collection strategy. They should define tasks for which the data is suitable and establish common evaluation metrics \cite{Lines2022}.

With a focus on practical applications, it is moreover crucial to consider the trade-offs involved in choosing one modality over another: ALS data is not as commonly available as VHR aerial imagery. Furthermore, the processing of point clouds presents additional challenges due to their large volume, unstructured nature, and a less mature deep learning landscape. In an ideal scenario, when considering ALS as a modality for tree species mapping, it should be compared to the more wildly available VHR aerial imagery and/or satellite imagery. Therefore, a good benchmark ALS dataset should be multimodal and present good alternative modalities to ALS.

\vspace{0.2cm}
In this paper, we make the following contributions: 
\begin{itemize}
\item Present PureForest: the largest publicly available dataset for tree species classification from ALS point clouds and VHR aerial images.
\item Set a baseline classification performance on PureForest from both Lidar and image modalities. 
\end{itemize}

We are inspired by a well-known public forest dataset: TreeSatAI \cite{treesatai}. TreeSatAI is a benchmark dataset for the classification of 15 tree species. It is a multi-modal dataset that includes both VHR aerial imagery and monotemporal acquisitions of Sentinel 1 and 2 satellite imagery. It consists of 50,381 60 m x 60 m patches distributed over a single state in Germany. We see TreeSatAI as a step towards large, diverse and representative tree species classification datasets that support deep learning research for forest monitoring. Our goal with PureForest is to provide a benchmark dataset with similar ambitions using ALS data.

\section{\justifying \textbf{Related Works}}\label{related_works}
\vspace{-0.2cm}
\subsection{Private Lidar datasets for tree species classification}

Deep learning studies on tree species classification from Lidar typically rely on ad-hoc, unpublished datasets, preventing reproducibility and comparison of methods. While the privacy of datasets is a common issue in most deep learning works related to forest monitoring \cite{Lines2022}, it is particularly striking for Lidar data, whose annotation is labor-intensive. Consequently, these private Lidar datasets typically have low spatial and species diversity, often containing a single forest with 3 to 5 semantic classes. Thus, studies based on these datasets may demonstrate some ability to classify a handful of tree species with good accuracy, but the generality of their conclusions is unclear. We illustrate these shortcomings in Table \ref{tab:datasets}, in which we report the characteristics of a subset of private Lidar datasets used in such studies.

Overall, this data-poor context leads to low reproducibility, higher research costs, and an inability to build on the work of others. 
 
\subsection{Public Lidar datasets for tree species classification}

As of February 2024, OpenForest \cite{openforest}, a regularly updated data catalogue of open forest datasets, only lists 3 Lidar datasets with tree species labels, which we report in Table \ref{tab:datasets}. Each of them covers at most 1500 trees, which gives them limited representativity and reduces their potential for model development and comparison.

This data gap motivated us to create PureForest, the largest ALS tree species classification dataset to date. PureForest is several orders of magnitude larger than existing public Lidar dataset for semantic segmentation, and has a larger number of semantic classes than most. Strikingly PureForest is 580 times larger than pytreedb \cite{weiser2022_dataset}, and 37 times more spatially diverse, with 449 distinct forests in PureForest versus only 12 in pytreedb.

\begin{table*}
\begin{minipage}{\textwidth}

\small
\centering
\setlength{\tabcolsep}{8pt}
\renewcommand{\arraystretch}{1.4}
\begin{tabular}{lcccccccc}
\textbf{Reference} & \textbf{Open access} & \textbf{Year} & \textbf{Lidar} & \textbf{Resolution (pts/m²)} & \textbf{Tree species labels} & \textbf{Area (ha)} & \textbf{Trees}   \\ \hline
\cite{Briechle2020} & \textcolor{red}{\ding{55}} & 2020 & ULS & 53 & 3 (+ dead trees) & 37 & -  \\
\cite{Liu2021}& \textcolor{red}{\ding{55}}  & 2021 & ULS & 40 & 2 & - & 1790  \\
\cite{Lv2022} & \textcolor{red}{\ding{55}} & 2022 & ULS & 40 & 4 & - & 2000 \\
\cite{Liu2022} & \textcolor{red}{\ding{55}}  & 2022 & BLS & - & 7 & 64 & -  \\
IDTReeS \cite{itc_dataset} & \textcolor{green}{\ding{52}}  & 2021 & ALS & 5 & 9 & 0.344 & - \\
Montmorency \cite{montmorency} & \textcolor{green}{\ding{52}}  & 2020 & RLS & - & 18 & 1.4 & 1453  \\
pytreedb \cite{weiser2022_dataset} & \textcolor{green}{\ding{52}}  & 2022 & ALS; ULS & 73; 1029 & 22 & 58.2 & 1491\\
\rowcolor[HTML]{e2e7f9}  \textbf{PureForest (ours)} & \textcolor{green}{\ding{52}}  & \textbf{2024} & \textbf{ALS} & \textbf{40} & \textbf{18 (in 13 classes)} & \textbf{33900} & \textbf{23.7M\footnote{Number of trees is calculated considering that there are 700 trees per hectare on average in metropolitan France according to a report from the French National Forest Inventory based on inventory data from 2018 to 2022 \cite{ifn_memento}}.}  
\end{tabular}

\caption{Overview of Lidar datasets suitable for tree species classification. We include publicly available Lidar datasets listed in the OpenForest catalog as of February 2024 \cite{openforest}. Furthermore, we list a few private datasets used in deep learning studies for tree species classification. ULS: Unmanned Lidar Scanning; ALS: Aerial Lidar Scanning; VHRAI: Very-High-Resolution Aerial Imagery; BLS: Backpack Lidar Scanning; RLS: Robot Lidar Scanning (i.e. unmaned ground vehicle).}
\label{tab:datasets}
\end{minipage}
\end{table*}

\section{\justifying \textbf{PureForest: The Dataset}}
We present PureForest, a benchmark dataset for tree species classification at the forest patch level. The dataset consists of 50 m $\times$ 50 m patches of ALS point clouds and aerial images, all annotated with a single semantic label. The annotations were generated in a semi-automated way and systematically validated by a trained forest expert from the aerial images. 

We present the data sources used to create PureForest, and describe the process used to create reliable tree species annotation polygon on a large scale. We then describe the content of the dataset, its class distribution and its structure. Finally, we list some of its limitations as well as potential uses. 

\subsection{Data sources for point clouds and aerial images}

The PureForest dataset features both 2D and 3D modalities, from two data sources: ALS data from the Lidar HD program and VHR aerial imagery from the ORTHO HR database. 

\textbf{Lidar HD}: The Lidar HD program (2020-2025) is an ambitious initiative undertaken by the IGN \cite{lidarhd}. The goal of the program is to obtain a 3D description of the French territory by 2025 using high-density ALS: 10 pulses/m², or about 40 pts/m². It is designed to meet the needs of public policies, such as flood risk prevention and estimation of forest resources. The data acquired and produced under this program are disseminated in open data (point clouds, DTM, DSM). Lidar~HD point clouds are classified, and we leave this classification unchanged in PureForest as it may facilitate ad hoc preprocessing such as height above ground normalization or extraction of tree features. The classes are others (value 1), ground (value 2), low vegetation (value 3), medium vegetation (value 4), high vegetation (value 5), building (value 6), water (value 9), bridge (value 17), permanent structures such as antennas and overhead power lines (value 64), artefacts points (value 65), and synthetic points (value 66). 

\textbf{ORTHO HR}: The ORTHO HR database is a mosaic of images acquired during national aerial surveys \cite{bdortho} by the IGN. Every year, about a third of French metropolitan territory is updated. The individual images are mapped onto a cartographic coordinate reference system and projected on a national DTM for orthorectification. The ORTHO HR imagery has a high spatial resolution of 0.20 m, and has near-infrared, red, green, and blue channels. Radiometric processing methods, including equalization and global correction, are applied to obtain the final product. However, the radiometry of channels cannot be considered a physical measurement of channel reflectance due to specific radiometric corrections applied for visual appealing. 

A key feature of PureForest lies in its multimodality, but the temporal synchronization of the two modalities is modest. The aerial imagery is acquired every three years for one third of the French territory, and the most recent acquisitions are used to curate PureForest annotations. In contrast, the ALS data in PureForest were collected from 2020 to 2023, resulting in a variable time lag of up to 3 years. Due to our inability to verify the label of every Lidar data patch, we must make the reasonable assumption that a complete change in tree species occurs only marginally in this time frame.

\subsection{Data sources for polygon annotation}

We adopt a semi-automatic annotation process, by first gathering polygons of likely pure forest areas, which are then validated and potentially corrected by a forest photointerpreter from recent aerial images. The polygons are obtained by crossing two forest databases produced by the IGN.

\textbf{\textit{BD Forêt}\textregistered V2}: The \textit{BD Forêt V2} is a forest vector database of tree species occupation in France, whose produced between 2007 and 2018 \cite{bdforet}. It is made of medium to large polygons (at least 0.5 ha). The polygons are annotated with labels of limited precision: if a species accounts for at least 75\% of an area, the forest polygon is labeled as “pure” for this species. While the \textit{BD Forêt V2} covers the full extent of metropolitan French forests, its definition of "pure forest” is not precise enough to benchmark and evaluate tree species classification methods.  

\textbf{French National Forest Inventory}: To reach higher levels of forest purity, we use ground truth data from the French National Forest Inventory (NFI) \cite{ifn_data}. The NFI conducts a statistical survey of French forests. Through photointerpretation and field surveys, dendrometric and ecological data are collected that provide valuable insights into the state, evolution and potential of French forests. Each year, these national surveys are used to produce reference publications on the French forest.

Our basic hypothesis is that forest plots with a single tree species are a strong (albeit highly local) signal that neighboring forests might also be monospecific. By taking this into account, we go beyond photoidentification and look for forests that are pure in a three-dimensional sense. This is important because ALS data can penetrate below the canopy, providing a signal that cannot be accounted for by photoidentification alone. 

\subsection{Annotation and data preparation}

\begin{figure}
\centering
\includegraphics[width=0.5\textwidth]{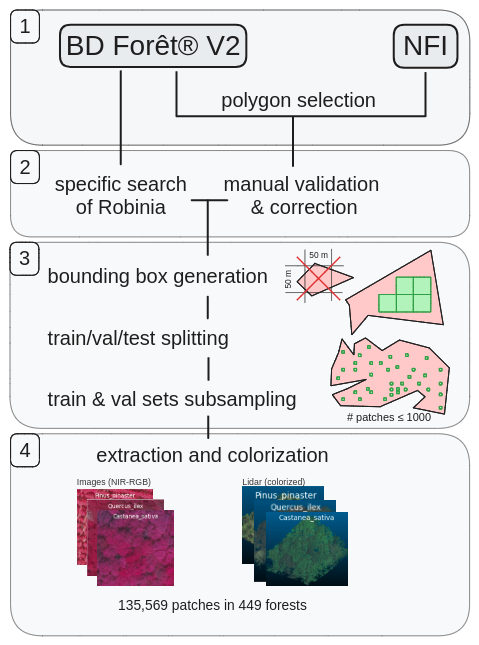}
\caption{Simplified flowchart of the annotation strategy. NFI = French National Forest Inventory.}
\label{fig:flowchart}
\end{figure}

Figure \ref{fig:flowchart} illustrates the multiple steps needed to turn large areas of likely pure forest into single-labeled data patches. 

\subsubsection{Polygon generation from reference databases}

At the time of dataset creation, Lidar HD data were only available for one third of the French metropolitan area, mostly in the south of France. In this area, 2756 forest plots were considered. Among them, 778 plots were monospecific according to the field surveys. In their vicinity, we selected \textit{BD Forêt V2} polygons labeled as pure (i.e. 75\% purity according to the definition mentioned above), closed forest, which gave us 569 polygons for a total area of 1974 km².

\subsubsection{Curation and additional photoidentification from aerial imagery}

The \textit{BD Forêt V2} was produced from aerial images from the years 2007-2018. However, to ensure synchronicity with the Lidar HD data, the annotation must take into account the most recent images available. Therefore, a trained operator verified each polygon, using aerial imagery from the Geoportail web platform within a standard GIS (QGIS). Photoidentification involved validating labels and editing polygon boundaries if there were discrepancies between the imagery and the polygons from the \textit{BD Forêt V2}. 534 validated polygons remained, covering a total area of 1018 km². 

In addition, pure inventory plots of Robina pseudoacacia were underrepresented. As they are relatively easy to identify, a specific search of polygons was performed by the photointerpreter in the \textit{BD Forêt V2}. This added another 133 annotation polygons for a total of 16 km² of Robinia forest. 

\subsubsection{Preparation of a deep-learning ready dataset}

\textbf{Train/val/test split}: To define a common benchmark, we divided the data into train, val, and test sets, with a stratification on semantic labels. The annotation polygons are scattered throughout the south of France, leading to a good geographical diversity within each semantic class. To account for spatial autocorrelation, the split is performed at the annotation polygon level: each forest belongs exclusively to either the train, val or test set.  This makes PureForest suitable for evaluating the territorial generalization of classification models. We aimed for a split of 70\%-15\%-15\% between the train, val and test sets.

\textbf{Patching}: Annotation polygons were divided by a regular 50 m $\times$ 50 m grid of square patches. Patches not completely contained within their annotation polygon were discarded to ensure that we only included monospecific forest patches. This operation filters out patches that overlap a polygon's boundaries and thus discards the smallest annotation polygons i.e. polygons that do not fit into a 50 m $\times$ 50 m patch. After this step, 449 annotation polygons remain.

\textbf{Subsampling}: In the train and val set only, patches in large annotation polygons were subsampled to limit their number to a maximum of 1000 patches per polygon. This subsampling reduces spatial redundancy and the disproportionate influence of large individual polygons, while also helping to balance class representation. It halved the total area of the dataset, resulting in a final size of 135,569 patches for an area of 339 km² of usable data. Note that we decided to maintain the original size of the test set in order to be able to study spatial patterns of error when evaluating models. As a result, the test set contains a larger number of patches (N=52935) than the train (N=69111) and val (N=13523) sets. The sizes of the sets can be found in Appendix \ref{appendix:trainvaltest}.

\subsubsection{Extraction of data patches}

Using the labeled geometric patches, we extracted the corresponding ALS point clouds and VHR aerial images. We then vertically colorized the Lidar data using near infrared, red, green and blue channels from the images, based on the vertical alignment of points and pixels.

\subsection{Dataset extent} 

\begin{figure}[h!]
\centering
\includegraphics[width=0.5\textwidth]{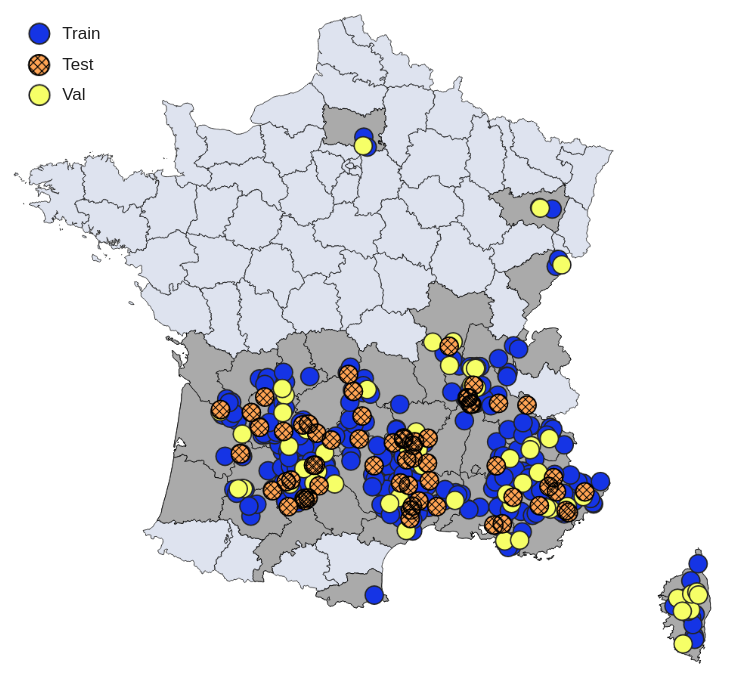}
\caption{Extent of the dataset in 40 department in metropolitan France, showing the spread of train (blue), val (yellow), and test (dashed orange) annotation polygons.}
\label{fig:extent}
\end{figure}

PureForest contains data from 449 distinct French forests, scattered across a large territory as determined by the random distribution of forest inventory plots. The forest are located in 40 French departments, which are depicted in dark grey in Figure \ref{fig:extent}. They are mainly located in the southern half of metropolitan France, reflecting the availability of Lidar~HD data at the time of dataset creation. 

\subsection{Semantic classes}

\begin{figure*}
 \centering
    \includegraphics[width=0.8\textwidth]{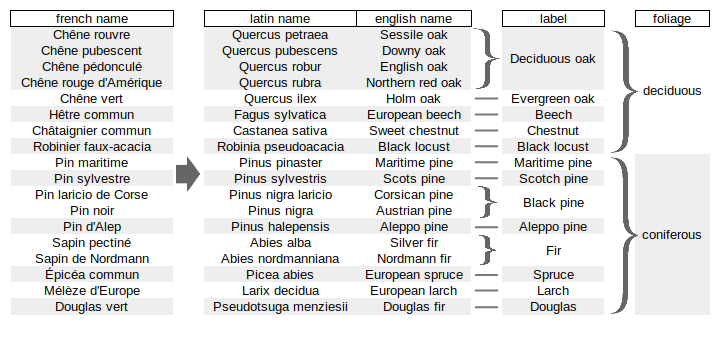}
    \caption{Semantic relationship between tree species, classification labels, and foliage type.}
 \label{fig:nomenclature}
\end{figure*}

Since we have access to accurate inventory ground truths, the annotations are first defined at the tree species level, resulting in 18 tree species (Figure \ref{fig:nomenclature}). Tree species are then hierarchically grouped into 13 semantic classes. This nomenclature comes from the \textit{BD Forêt V2}, with one difference: “Fir” and “Spruce” make two distinct classes instead of being grouped together.

\begin{table}[htpb]
\small
\centering
\setlength{\tabcolsep}{4.9pt}
\renewcommand{\arraystretch}{1.6}
\begin{tabular}{p{1.9cm}lcrrrr}
\textbf{Class}              & \textbf{ID}       & \textbf{Patches}   & \textbf{\%} & \textbf{Polygons} & \textbf{Area (km²)} \\ \hline
Deciduous oak           & 0                & 48055 & 35.4& 91	& 120.14       \\
Evergreen oak   & 1                & 22361 & 16.5& 36	& 55.9       \\
Beech & 2                & 12670 & 9.3& 29	& 31.68      \\
Chestnut          & 3                & 3684   & 2.7 & 21	& 9.21      \\
Black locust              & 4                & 2303 & 1.7 & 107	& 5.76      \\
Maritime pine         & 5                & 7568   & 5.6 & 27	& 18.92      \\
Scotch pine          & 6                & 18265 & 13.5 & 46	& 45.66     \\
Black pine          & 7                & 7226 & 5.3 & 22	& 18.07      \\
Aleppo pine           & 8                & 4699   & 3.5  & 19	& 11.75     \\
Fir          & 9   & 840 & 0.6 & 4	& 2.1     \\
Spruce               & 10  & 4074 & 3.0 & 23	& 10.18     \\
Larch        & 11               & 3294   & 2.4 & 9	& 8.24     \\
Douglas     & 12 & 530   & 0.4 & 15	& 1.32      \\ \hline
\end{tabular}
\caption{Support of each semantic class in terms of patches, distinct annotation polygons, and area covered by the patches.}
\label{tab:supports}
\end{table}

Train/val/test splitting happens at the annotation polygon level, and those can have very different sizes. Even though limiting to 1000 patches by polygon reduces the disproportionate importance of the largest polygons, prevalence of each class (in terms of number of patches) might still vary between splits. This is shown in the histograms in Figure \ref{fig:class_distribution}, and the numbers can be found in \ref{appendix:supports}. In the test set, all classes are represented with at least 100 patches, with the notable exception of Fir, represented with only 22 patches from a single test polygon. 

 \begin{figure}
     \centering
        \includegraphics[width=0.5\textwidth]{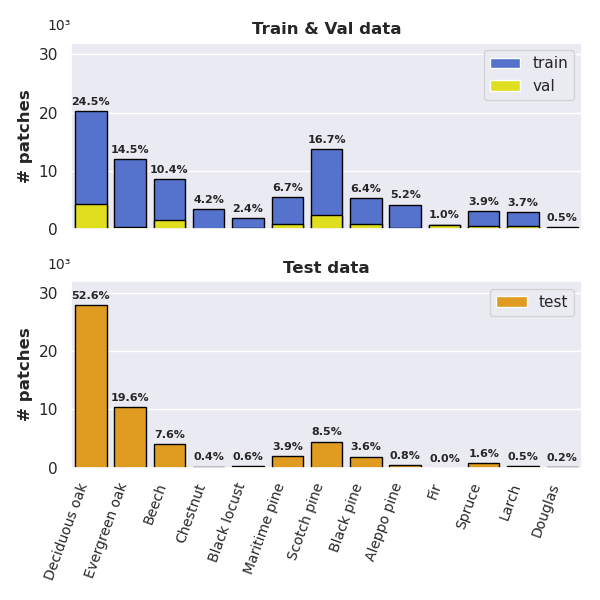}   
        \caption{Class distribution of the train and val set (top) and test set (bottom). Samples are 50 m $\times$ 50 m patches. A thousand patches make 2.5 km².}
     \label{fig:class_distribution}
 \end{figure}

Our annotation methodology involves the use of NFI plots, which leads to stochastic sampling of annotation areas and therefore to stochastic variations in class frequencies, especially for rarer tree species. Furthermore, we expect that tree species that are mostly found in mixed forests will be underrepresented. Finally, we expect a bias towards southern tree species, as southern France is overrepresented in PureForest due to Lidar~HD data not being acquired yet in most northern regions at the time of dataset creation.

Thus, the dataset suffers from class imbalance, with class supports ranging from 530 patches (0.4\%) for class Douglas to 48,055 patches (35.4\%) for class Deciduous Oak (Table \ref{tab:supports}). Out of the 13 classes, only 2 of them account for less than 5 km² of exploitable data: Fir (2.1 km²) and Douglas (1.32 km²).

\subsection{Dataset structure}

\begin{figure}[h!]
\centering
\renewcommand{\arraystretch}{1.5}

\par\vskip1.2pt
\hspace{0.02cm}
\begin{Tabular}[0.8]{|p{8.2cm}|}
\hline 
\rowcolor[HTML]{f7f9fd}  {\footnotesize
\begin{forest}
  pic dir tree, where level=0{}{directory,},
  for tree={ s sep=0.05cm, l sep=0.65cm, font=\rmfamily }
  [\textbf{PureForest}
    [\textbf{lidar}
      [\textbf{train\textbackslash val\textbackslash test}
        [\texttt{\{SPLIT\}-\{species\}-C\{class\_id\}-\{id\}}.laz, file]
      ]
    ]
    [\textbf{imagery}
      [\textbf{train\textbackslash val\textbackslash test}
        [\texttt{\{SPLIT\}-\{species\}-C\{class\_id\}-\{id\}}.tiff, file]
      ]
    ]
    [\textbf{metadata}
      [PureForest-patches.csv, file]
      [PureForest-patches.gpkg, file]
      [PureForest-dictionnary.csv, file]
      ]
  ]
\end{forest}}      \\ \hline
\end{Tabular}
\vfill
\caption{Structure of files and directories in PureForest}
\label{fig:dataset_structure}
\end{figure}
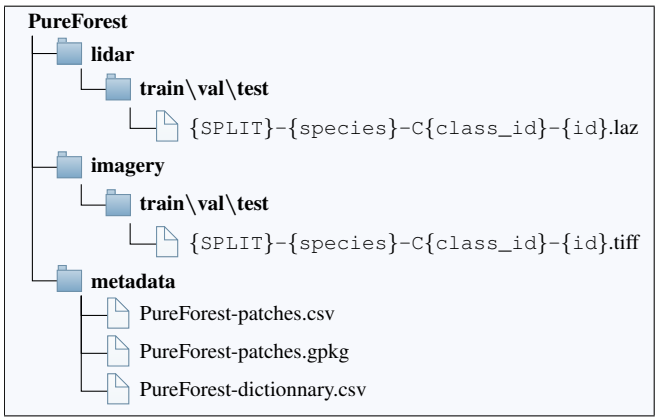

Figure \ref{fig:dataset_structure} illustrates the dataset file structure.

Directory \texttt{imagery} contains the VHR aerial images patches derived from most recent ORTHO HR imagery data at the time of photoidentification (January 2023). Patches size is 50$\times$50 m with a 0.2 m spatial resolution (250$\times$250 pixels). Band order is near-infrared, red, green, blue.

Directory \texttt{lidar} contains the corresponding ALS point clouds from Lidar HD data (2020-2025), in LAZ 1.4 format. Point density is high, with approximately 40 points/m². Clouds are colorized by their corresponding aerial image. 

\begin{simplechar}
Naming convention for data patches is shown in Figure \ref{fig:dataset_structure}, where \texttt{SPLIT} is either TRAIN, VAL, or TEST, \texttt{class_id} indicates the semantic class identifier (zero-based), and \texttt{patch_id} is a unique patch identifier. For instance, \texttt{TEST-Quercus_pubescens-C0-199_7_327.laz} refers to a Lidar point cloud of a deciduous oak forest from the test set. 
\end{simplechar}

In folder metadata, one can find three files:
\vspace{-0.75em}
\begin{itemize}[leftmargin=1em]
    \item \texttt{PureForest-patches.gpkg}: listing of all patches with class labels and some metadata, including their membership to an annotation polygon.
    \item \texttt{PureForest-patches.csv}: same content, except without patch geometries, provided for convenience.
    \item \texttt{PureForest-dictionnary.csv}: gives a reference mapping for all species present in the dataset to their french/english/latin names, their category in the \textit{BD Forêt V2}, and their species code as defined by the NFI.
\end{itemize}

\subsection{Methodological limits}
\begin{itemize}[leftmargin=1em]

\item Only 40 French administrative departments are considered in PureForest due to the limited availability of Lidar HD data at the time of creation. While this represents a significant spatial extent, it does not fully capture the diversity of French forest presentation for the 13 considered semantic classes. These departments being primarily located in Southern France, the dataset is slightly biased towards higher concentrations of southern tree species. 

\item Our semi-automated annotation approach allowed us to generate a large number of annotations with minimal human labor. However, it is important to note that each tree species is only represented by a limited number of distinct forests, with PureForest having an average of 35 annotation polygons (about 26 km² of area) per tree species. This should be taken into consideration when studying the ability of a tree species classification model to generalize. 

\item While we ensured that the annotation polygons in the val and test sets were different from those in the train set, we did not enforce a spatial buffer between polygons in different sets, and thus did not explicitly control for spatial autocorrelation between them. Note that out of 449 annotation polygons, we observe only a handful of cases where a train polygon is less than 1 km away from a val or test polygon of the same species. 

\item We focused on monospecific, closed forests based on the labels of the \textit{BD Forêt V2}. This simplifies both annotation and model benchmarking. However, extending the data to mixed and/or open forests would be necessary to create a national-scale forest model. This could be achieved through the use of multi-label annotations or individual tree labelling.
\end{itemize}

\subsection{Possible uses for PureForest} 

From a broad perspective, PureForest can be used for the study of forests and for the development of technical methodologies that rely on the processing of either ALS point clouds or VHR aerial images to monitor them. More specifically, PureForest is an interesting benchmark dataset for the development of neural architectures and deep learning methods, whether general-purpose or designed around the specific characteristics of forest data. Regarding multimodality, there is potential to explore 3D-2D modality fusion beyond Lidar colorization. 

PureForest is primarily suited for tree species classification at the forest patch level thanks to its coverage (339 km²), spatial diversity (40 French departments), semantic diversity (13 classes), and the high resolution of the ALS data (40 points/m²).

A related task is the classification of tree species at the single-tree level. Preprocessing the dataset using state-of-the-art individual tree crown segmentation (ITCS) algorithms and making the results publicly available could greatly benefit further research in this area. In terms of individual tree crown segmentation, it is important to evaluate how well existing methods for ITCS generalize to different tree species. This evaluation would require additional labeling work. 

Given the high volume of data available, there is an opportunity to develop semi-supervised learning approaches that leverage both labeled and unlabeled data for improved model performance.

\begin{figure*}
 \centering
    \includegraphics[width=1.0\textwidth]{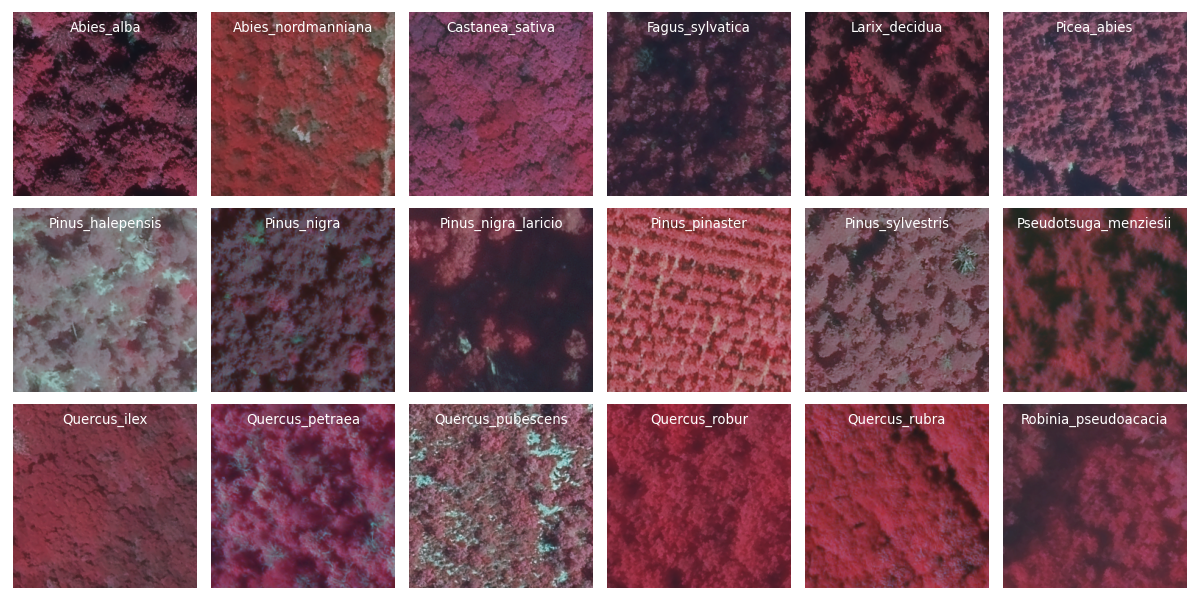}
    \includegraphics[width=1.0\textwidth]{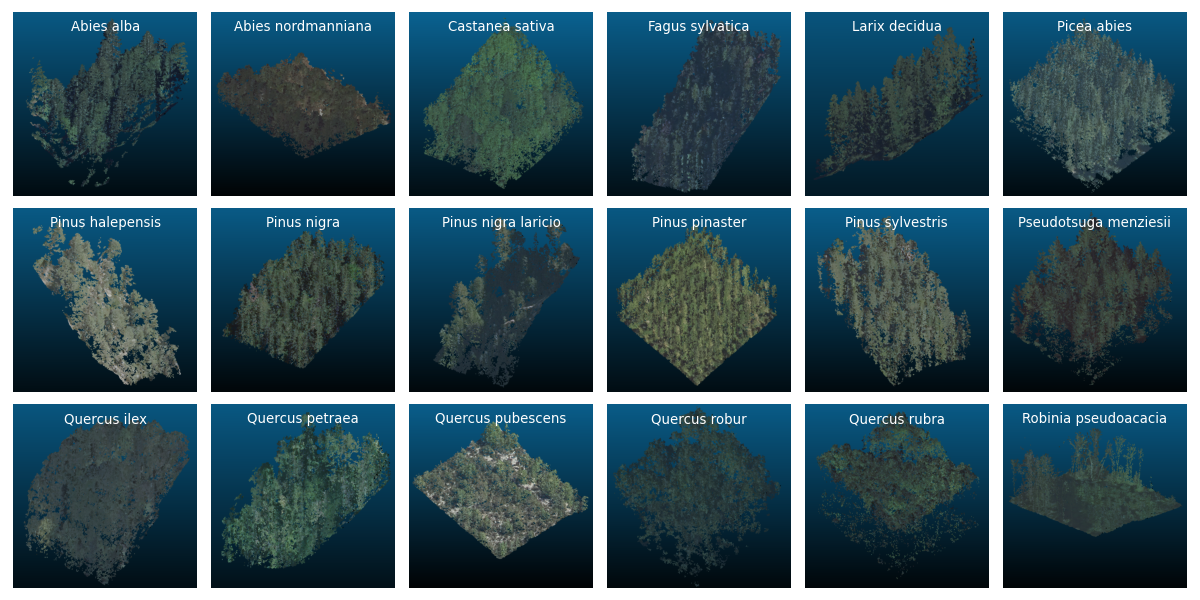}
    \caption{Point cloud and image samples for each of the 18 tree species in PureForest. Aerial images are displayed in fake colors (near-infrared, red, green)}
 \label{fig:patches_viz}
\end{figure*}

\section{\justifying\textbf{Benchmark models}}
To give an idea of the achievable performance on PureForest, we establish a baseline classification performance from the Lidar modality. 
While advanced multimodal modeling is beyond the scope of this paper; we also investigate two possible extensions of the baseline model. First, we evaluate the impact of considering colorized point clouds rather than relying solely on the geometric and reflectance information of ALS data. Second, we evaluate the impact of integrating contextual metadata, specifically the elevation (above sea level) of a forest. Finally, and in order to demonstrate how ALS compares to the more commonly used VHR imagery for the task at hand, we establish a baseline classification performance from VHR aerial images.

\subsection{Overview of 3D Deep Learning architectures}\label{architectures}

To provide some background, we give a brief overview of 3D deep learning approaches, and then talk briefly about their application to tree species classification.

Processing unstructured, unordered sets of points is challenging, and researchers would initially try to turn point clouds into structured data i.e. put them “on a grid”. In voxel-based models such as SegCloud \cite{segcloud}, 3D convolutions process a voxelized version of a point cloud. In multi-view approaches like SnapNet \cite{snapnet}, 2D views or projections of point clouds would be processed by traditional 2D convolutional neural networks. In these methods, the semantic segmentation happens on a regular grid and is then projected back to the point cloud. 

Deep learning that operates directly from point clouds is recent \cite{3dsurvey}. In 2016, \cite{pointnet} introduce PointNet, a pioneering architecture that deals directly with disordered point sets using multiple shared MLPs and symmetric pooling operations. One year later, \cite{pointnet2} propose PointNet++, which builds on PointNet layers to process points clouds on nested partitions of the input point cloud. PointNet++ is the best known representative of point-based methods which are characterized by PointNet-like operations hierarchically organized in a U-shaped architecture. From a performance point of view, an interesting successor of PointNet++ is RandLA-Net \cite{randlanet}. Introduced in 2019, RandLa-Net features some performance improvements thanks to the use of random subsampling combined with an explicit consideration of the relative positions of points in space. This makes it a suitable architecture for the large scales characteristic of remote sensing applications.

Point clouds can be represented as graphs: nodes correspond to points and edges capture spatial relationships, which can then be processed by a Graph Convolutional Networks (GNN). In 2018, DGCNN (Dynamic Graph CNN) was introduced by \cite{dgcnn}, in which edge convolutional layers capture geometric structures. Alternatively, Graph Attention Networks (GAN) have been proposed: in 2021, \cite{pct} introduce the Point Cloud Transformer, which fully relies on attention layers to capture geometric structures. 

Along 3D neural architectures, new representations of point clouds have been proposed. In 2017, \cite{spg} introduce their SuperPoint Graph, a partitioning of a point cloud scene into groups of semantically homogeneous points that can be processed with high efficiency by graph neural networks. Most recently, \cite{spt} turned the SuperPoint Graph into a Hierarchical SuperPoint Partition i.e. a partition of nested, increasingly coarser SuperPoints. Additionally, they propose a variant of graph-attention networks to operate directly on the SuperPoint Partition, resulting in highly efficient cloud processing.

For tree species classification from Lidar point clouds, PointNet++ is considered by many as the de facto solution (\cite{Briechle2020}, \cite{Lv2022}, \cite{Liu2021}, \cite{Liu2022}). While not the most recent architecture, it is still highly robust and competitive. Interestingly, \cite{Liu2022} show that performance is task dependent. Compared to 5 other state-of-the-art architectures, PointNet++ underperforms in a generic 3D object classification benchmark but is among the top three performers in classifying 7 tree species from Unmanned Laser Scanning (ULS) data. In addition, we note that some researchers choose to transpose point cloud data into the 2D domain, allowing them to use image-based methods. For example, \cite{zhong2024} build a canopy height model (CHM) from ULS data, which they then feed into a pre-trained image neural network.

\subsection{Baseline: classification from Lidar point clouds}\label{3d_baseline}

The benchmarking code is based on an existing 3D deep learning library: Myria3D \cite{myria3d}. Myria3D is developed under the Pytorch-Lightning framework \cite{lightning}, featuring Pytorch-Geometric \cite{pyg}, a powerful framework for 3D deep learning. It was explicitly designed for the semantic segmentation of Lidar HD data. In our experiments, we kept most defaults setting from Myria3D, including choice of architecture, data processing methods, and training hyperparameters.

\subsubsection{A 3D neural architecture for Lidar scene classification}

\begin{figure*}
\centering
\includegraphics[width=\textwidth]{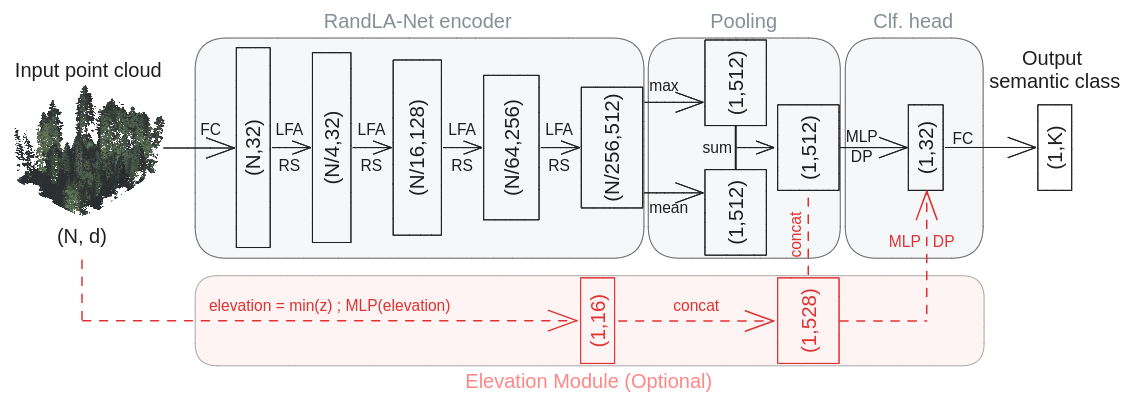}
\caption{The detailed architecture of the truncated (encoder-only) RandLA-Net, which we adapted for multiclass classification. $(N, D)$ represents the number of points and number of features, respectively. Input dimensions is $d$ and $k$ is the number of semantic classes. The path for the optional integration of elevation is in \textcolor{red}{red}. Clf. head: Classification Head, FC: Fully Connected layer, LFA: Local Feature Aggregation, RS: Random Sampling, MLP: shared Multi-Layer Perceptron, DP: Dropout.}
\label{fig:architecture}
\end{figure*}

As a benchmark architecture, we use RandLA-Net \cite{randlanet}, mentioned in Subsection \ref{architectures}. RandLA-Net was originally designed for semantic segmentation and we adapt it for scene classification by removing its decoder, thus simplifying the architecture. Summing the max-pool and mean-pool results of activations at the lowest resolution gives us a point cloud embedding. This embedding is then fed to a simple 2-layer multi-layer perceptron (MLP) followed by a fully connected (FC) layer to obtain class logits. Figure \ref{fig:architecture} illustrates this truncated architecture.

\subsubsection{Preprocessing of point clouds}

Lidar is unstructured and how we feed clouds to a model is important and has a lot of degrees of freedom. We first subsample clouds via a grid with a voxel size of 0.25 m. A maximal budget of 40,000 points per cloud is allocated, enforced when needed via a random subsampling. Clouds are horizontally centered by subtracting their average position along the x and y axes, and vertically aligned by subtracting their minimal position along the z axis. Centered coordinates are then divided by 25 (meters) to be homogeneously brought between –1 and 1.

In terms of features, the following Lidar dimensions are included: x, y, z, reflectance, echo number, number of echos. Number of echos and echo number are normalized by constants to be between 0 and 1. Reflectance is log-normalized, standardized, and values above three standard deviations are clamped. Note that at this stage, and to define a baseline model focusing solely on the Lidar modality, we do not consider color information. 

Finally, we apply several generic data augmentations, namely: Random Scale (x0.8 to x1.25 factor), Random Jitter (-0.05 m to +0.05 m along each axis), Random Translate (-1 m to +1 m), Random Flip (along x and y axes).

\subsubsection{Hyperparameters}

Training is supervised with Cross-Entropy loss and the Adam optimizer. The learning rate is 0.004, with a reduction strategy (ReduceLROnPlateau) that halves the learning rate with a patience of 20 epochs and a cooldown of 5 epochs after each reduction. A batch size of 10 is used. We train the model for at least 100 epochs and retain the model that minimizes the validation loss.

\subsubsection{Infrastructure}

Training is conducted using 3 NVIDIA Tesla V100 GPUs, each equipped with 32 GB of memory, housed within an in-house High-Performance Computing (HPC) cluster. PyTorch-Lightning's distributed data parallel (ddp) strategy is employed to leverage multi-GPU data distribution. Training this particular implementation of RandLA-Net under these settings typically spans about 24 hours. 

\subsection{First extension of Baseline: colorized Lidar}

Alongside contextual information, photointerpreters utilize the spectral information of aerial imagery to discriminate tree species. For instance, there is a large spectral difference between coniferous and deciduous trees in the near infrared spectrum \cite{photointerpretation}. Lidar colorization is the most straightforward way to consider spectral information. All things being equal, we update the baseline model to consider color information, namely the near infrared, red, green, blue dimensions of the colorized point clouds. We set all colors to 0 for points with an echo number above 1, as a very basic occlusion model. Furthermore, we compute the Normalized Difference Vegetation Index (NDVI). We also calculate a new “color intensity” feature as the average of red, green, and blue channels. All colors are divided by a constant to be between 0 and 1. Average color is normalized similarly to the reflectance feature i.e. by log-normalization, standardization, and truncation of values above three standard deviations.

\subsection{Second extension of the Baseline: contextualization with patch elevation}

Elevation provides information on which tree species can thrive in a given forest, as certain species are exclusive to specific elevation ranges . While elevation is available in public national DTMs, it is also readily available in Lidar data, which makes it a valuable metadata with a low access and development cost. We first calculate elevation as the minimal z value within a patch. To avoid overfitting on the limited set of elevations, we randomly shift elevations by 100 meters maximum. We then normalize the feature to be between 0 and 1. Finally, the normalized elevation is embedded by a small MLP with dimensions [1;16;16], whose output is concatenated with the cloud’s embedding obtained from pooling the post-encoder activations. Figure \ref{fig:architecture} illustrates this optional integration of elevation in red. 

\subsection{Classification from aerial images}

VHR aerial imagery is more commonly available than ALS data, and it is important to evaluate how the two modalities compare. Therefore, we also train and evaluate a basic image classifier on aerial imagery. We adopt the experimental context reported by creators of TreeSatAI for their own baseline \cite{treesatai}: a ResNet18 \cite{resnet} encoder with a batch size of 32, optimized with Adam with a cyclic learning rate (0.00005-0.001 range, half-period of 13630 steps). The model is pretrained on ImageNet and we adapt data processing accordingly: images are resized to 224 $\times$ 224 pixels, the model is adapted to receive the additional infrared channel by duplicating the weights for the red channel, and color channels are normalized using ImageNet statistics. We use common data augmentations: random rotations, horizontal and vertical flips, color jitter (p=0.5), random cropping (p=0.5), channel dropout (p=0.05).

Training is parallelized as before with 3 GPUS, and we scale the learning rate range linearly to account for the effective batch size. Since the model is pretrained, it converges in a few training epochs and then quickly plateaus. This narrows the window of good models, and therefore we validate the model more frequently, i.e., every tenth of the training set. Training takes 4 minutes per epoch on average.

\subsection{Task definition and model evaluation}
PureForest is designed for a classification task in which the goal is to discriminate between 13 classes of tree species in pure forests settings.

To evaluate classification approaches, we first define a confusion matrix calculated between the labels and the model predictions. Subsequently, we calculate the Overall Accuracy (OA). Since the data is highly imbalanced, we also calculate the weighted variants of precision, recall, and F1-score. Due to the spatial nature of the data, classification of patches is akin to the semantic segmentation of large 50 m x 50 m pixels. Therefore, we decide to also compute the Intersection-over-Union (IoU, also called Jaccard index), which is a well-behaved classification metric that is robust to class imbalance and is typically used for semantic segmentation tasks. It is computed for each class with the following formula: 

\begin{equation*}
    IoU = \frac{TP}{TP+FP+FN} 
\end{equation*}\vskip 0.05cm

where $TP$ = true positives, $FP$ = false positives, and $FN$ = false negatives. As a global metric to evaluate and rank models, the mean IoU (mIoU) is defined as the average of the per-class IoU. mIoU ensures equal consideration for each class regardless of its dataset frequency.

\section{\justifying\textbf{Experimental Results}}
In Table \ref{tab:results_models} we report for all models the results obtained using the 69,111 patches for training and 13,523 patches for validation, and testing on the remaining 52,935 patches of PureForest.

\begin{table*}[htpb]
\small
\centering
\setlength{\tabcolsep}{12pt}
\renewcommand{\arraystretch}{1.6}
\begin{tabular}{lcccccc}
& & & Weighted & & & \\ \cline{3-5}
\textbf{Modality} & \textbf{Epoch saved} & \textbf{Precision} & \textbf{Recall} & \textbf{F1-Score} & \textbf{OA} & \textbf{mIoU} \\ \hline
\rowcolor[HTML]{e2e7f9} Lidar (Baseline) & 26 & 86.3 & 80.3 & 81.0 & 80.3 & 55.1 \\  
Lidar + RGBI & 12 & 85.6 & 79.1 & 79.6 & 79.1 & 53.6 \\
Lidar + Elevation & 28 & 87.6  & 83.6 & 84.4 & 83.6  & 57.2 \\ \hline
Aerial Imagery & 1 & 81.9 & 73.1 & 74.6 & 73.1 & 50.0 \\ \hline
\end{tabular}
\caption{Performance metrics on the test set for the Baseline model based on Lidar data only, its extensions on colorized Lidar and with elevation, and the model based on aerial images only. OA=Overall Accuracy. Note: the results of the 3D and 2D models should be considered as baselines, and are not enough to conclude on the practical superiority of one modality over another.}
\label{tab:results_models}
\end{table*}

\subsection{Results for the baseline model}

Using only Lidar data, the baseline model performs well globally with an OA of 80\%. It also achieves good performance across most classes, with a mIoU of 55\% and weighted precision, recall and F1 score all above 80\%.

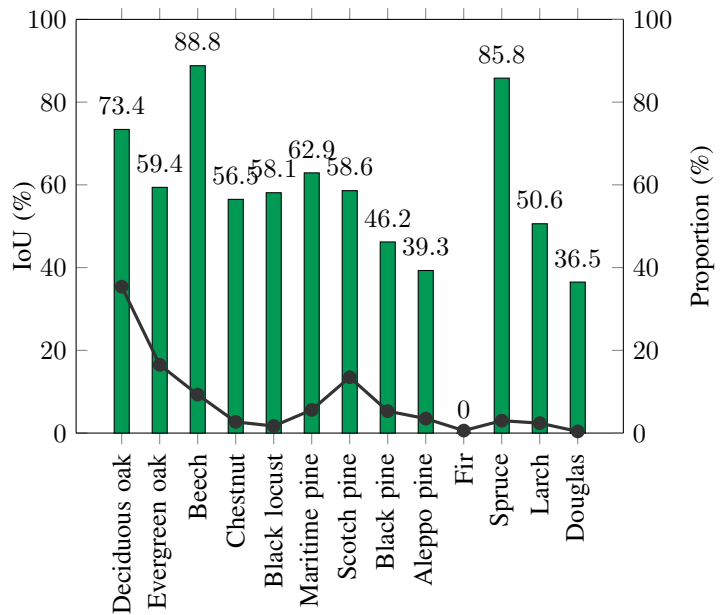
\begin{figure}
\begin{tikzpicture}

\begin{axis}[
    width=1\linewidth,
    height=.3\textheight,
    ybar,
    bar width=.2cm,
    axis y line*=left, %
    ylabel shift = -1 em,
    ymin=0,
    ymax=100,
    nodes near coords,
    nodes near coords style={shift={(0ex,0.5ex)}},
    nodes near coords align={vertical},
    symbolic x coords={Deciduous oak,Evergreen oak,Beech,Chestnut,Black locust,Maritime pine,Scotch pine,Black pine,Aleppo pine,Fir,Spruce,Larch,Douglas},
    legend style={at={(0.5,-0.50)},
      anchor=north,legend columns=-1},
    xtick=data,
    ylabel={IoU (\%)},
    xticklabel style={rotate=90,anchor=east},
    ylabel near ticks, 
    yticklabel pos=left,
]
\addplot[fill=ForestGreen, text=black] coordinates {(Deciduous oak, 73.4) (Evergreen oak, 59.4) (Beech, 88.8) (Chestnut, 56.5) (Black locust, 58.1) (Maritime pine, 62.9) (Scotch pine, 58.6) (Black pine, 46.2) (Aleppo pine, 39.3) (Fir, 0) (Spruce, 85.8) (Larch, 50.6) (Douglas, 36.5)};
\end{axis}

\begin{axis}[
    axis y line*=right,
    ylabel near ticks,
    yticklabel pos=right,
    ylabel={Proportion (\%)},
    symbolic x coords={Deciduous oak,Evergreen oak,Beech,Chestnut,Black locust,Maritime pine,Scotch pine,Black pine,Aleppo pine,Fir,Spruce,Larch,Douglas},
    xtick=data,
    xticklabel style={rotate=90,anchor=east},
    nodes near coords align={vertical},
    axis x line=none,
    ymin=0,
    ymax=100,
    width=1\linewidth,
    height=.3\textheight,
    bar width=.2cm,
    ]
\addplot [mark=*,very thick, black!80] coordinates {(Deciduous oak,35.4) (Evergreen oak,16.5) (Beech,9.3) (Chestnut,2.7) (Black locust,1.7) (Maritime pine,5.6) (Scotch pine,13.5) (Black pine,5.3) (Aleppo pine,3.5) (Fir,0.6) (Spruce,3.0) (Larch,2.4) (Douglas,0.4)};
\legend{}; %
\end{axis}

\end{tikzpicture}
\caption{Class IoUs for the Lidar baseline model (green bars), contrasted with class proportions in PureForest (black line).}
\label{fig:classwise_perfs_v2}
\end{figure}

Figure \ref{fig:classwise_perfs_v2} illustrates its classwise performance, in relation to the relative proportion of each class in PureForest. Coherent with the common intuition that more data leads to better model performances, class IoU is positively correlated with class frequency, with a Pearson's correlation coefficient of 0.41. We observe that 9 out of the 13 semantic classes have an IoU above 50\%. Classes with an IoU under this threshold are all conifers: Black Pine (IoU=46.2\%), Aleppo Pine (IoU=39.3\%), Douglas (IoU=36.5\%), and Fir (IoU=0.0\%). Considering that Fir is the rarest class in the train set with only 94 patches (0.14\% of the train set) this last result is not surprising. The best performing classes are Deciduous oak (IoU=73.4\%), Spruce (IoU=85.8\%) and Beech (IoU=88.8\%), and other classes have a IoU in the range of 50\%-65\%.

\begin{figure}
     \centering
        \includegraphics[width=0.5\textwidth]{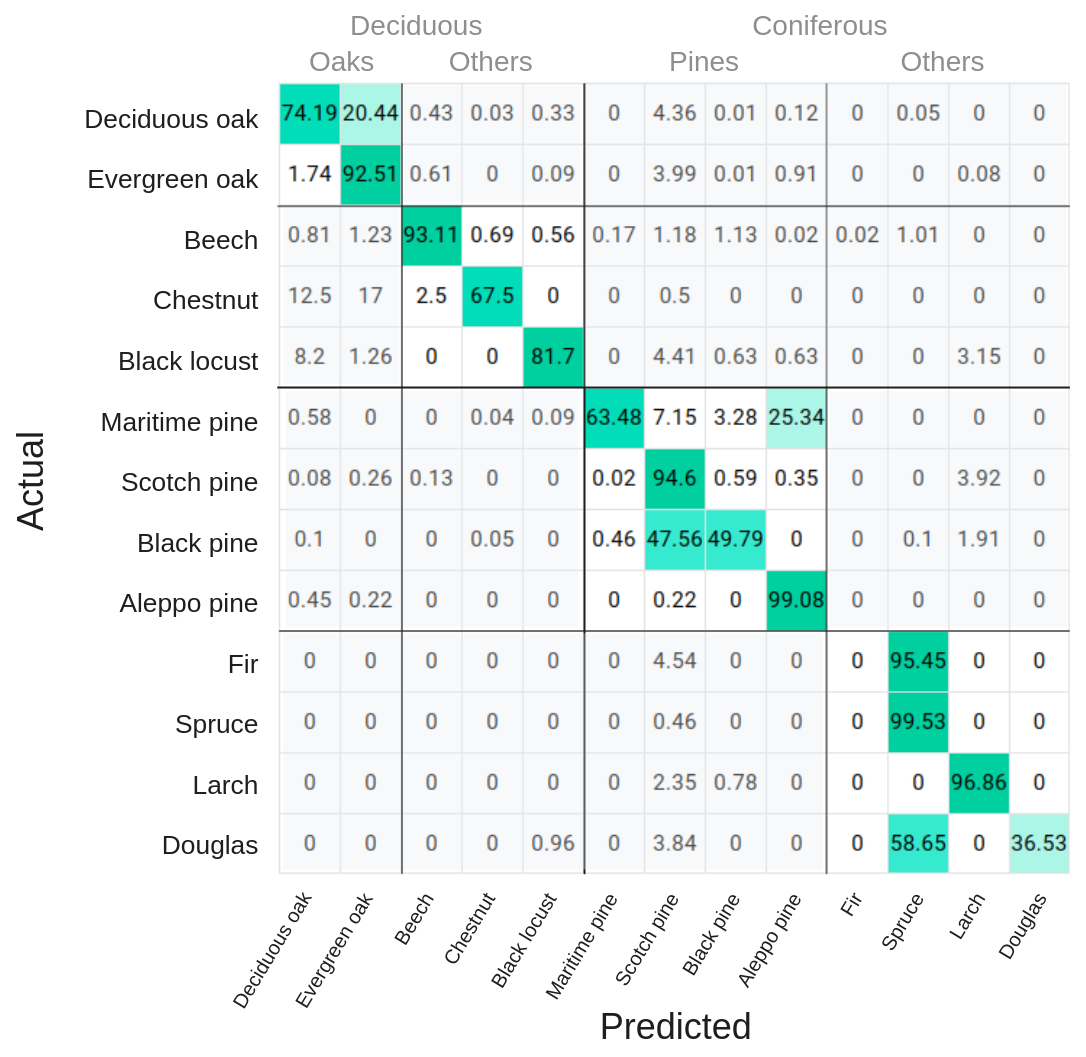}   
        \caption{Baseline confusion matrix on the test set normalized by rows. Hierarchical relationships between semantic classes are highlighted: Oaks and Others among deciduous trees, Pines and Others among coniferous trees}
     \label{fig:cm}
 \end{figure}

Figure \ref{fig:cm} provides the confusion matrix of the baseline Lidar model on the test set. Interclass confusions are mainly found between close tree species: Deciduous oak and Evergreen oak, Maritime Pine and Aleppo Pine, Scotch Pine and Black Pine, Fir and Spruce, Douglas and Spruce. One important source of errors is due to patches from all classes incorrectly predicted as Scotch Pine forests, with 2.8\% of patches in each non-Pine class concerned on average. This pattern may be related to Scotch pine being twice as frequent in the train and val sets (16.7\%) than in the test set (8.5\%). Interestingly, while the Lidar model fails on class Fir, it was able to capture its semantic proximity with Spruce despite Fir being extremely rare in the train set. Overall, these results demonstrate that the main semantic relationship between tree species are efficiently captured by 3D models from the Lidar data in PureForest.

\subsection{Results for other models}

Interestingly, the model based on colorized Lidar performs worse, with a small drop of up to 1.5\% for all global metrics. It also stops learning sooner: the validation loss stops decreasing after only 12 training epochs instead of 26 for the model processing raw Lidar. We hypothesise that colorization brings more noise than signal because a 3D model like RandLA-Net is not capable of exploiting the texture of aerial imagery, which is key for species classification. This limitations points toward the high relevance of exploring multi-modal approaches on PureForest, specifically those that are able to exploit both 2D textures and 3D structures.

The Lidar model benefits from the explicit integration of the patch elevation with an ad-hoc neural module and relevant data augmentation. While training reaches its optimum after a similar number of epochs, this simple contextualization adds 2\% of mIoU and 3\% of OA. While limited, this improvement illustrates the potential of integrating relevant contextual metadata like the type of soil, sunlight exposure (i.e. north-facing versus south-facing slopes), or even the historical distribution of tree species in the patch surrounding area, in a way that is architecture-agnostic.

In our basic experimental settings, the model trained on VHR aerial images from PureForest underperforms, with an mIoU 5\% points below the one of the baseline Lidar model. Interestingly, with a weighted F1-Score of 74.6\%, is is not far from the weighted F1-Score of 69.5\% achieved in TreeSatAI baselines (see Table 1 in \cite{treesatai}). However, we emphasize that neither hyperparameters nor model architectures were searched, and thus this result is not suitable to infer superiority of one modality over the other. It should be regarded as demonstration that ALS is a competitive modality for tree species classification, on par with the more commonly used VHR aerial imagery.

\section{\justifying\textbf{Conclusion}}
We presented a large-scale benchmark dataset covering 339 km² of monospecific forest with verified labels in the form of high-density ALS point clouds and VHR aerial images. PureForest is the most extensive publicly available Lidar dataset of its kind. It has both semantic and spatial diversity: its data come from 449 forests in 40 French administrative departments, with 18 tree species grouped into 13 semantic classes. It is also representative of French forests for the semantic classes considered. We established baseline classification performances and showed the potential of its 3D and 2D modalities. These evaluations showed that classification in PureForest is challenging enough for researchers to use it as a robust benchmark for scene classification in general and tree species classification in particular. Since the dataset is georeferenced, it can be augmented with contextual metadata such as soil types, or with other modalities such as satellite imagery. We hope that PureForest will be a resource for the research community and help advance the field of deep learning for Lidar.

\section*{\justifying\textbf{Code and data access}}
The PureForest dataset is hosted on HuggingFace under the Open Licence 2.0 of Etalab: \href{https://huggingface.co/datasets/IGNF/PureForest}{IGNF/PureForest}.

Benchmark code for 3D models comes from the Myria3D library which was adapted for the task of scene classification, as explained in Subsection \ref{3d_baseline}. Code is available on specific Github branches at \href{https://github.com/IGNF/myria3d}{github.com/IGNF/myria3d}.

Benchmark code for the image classifier is available at \href{https://github.com/IGNF/PureForest-Baseline}{github.com/IGNF/PureForest-Baseline}.

\section*{\justifying\textbf{Acknowledgements}}
The authors thank Daniel Mijalcevic for the verification and correction of annotation polygons, and Remi Pas and Michel Daab for their contributions to data engineering, and Anatol Garioud, Nicolas Gonthier, Antoine Labatie and Matthieu Porte for reviewing this data paper.

\section*{\justifying\textbf{Authors' contribution}}
C.G. and F.R defined the annotation methodology. C.G. implemented the tools for data engineering. C.G. implemented and evaluated the deep learning baseline. C.G. wrote the data paper and released the dataset.

\newpage

\section*{\justifying\textbf{References}}
\bibliographystyle{IEEEtran}
\bibliography{export}
\newpage
\onecolumn

\section*{\justifying\textbf{Appendices}}
\renewcommand{\thesubsection}{\textbf{\Alph{subsection}}}
\renewcommand\thesubsectiondis{\textbf{\Alph{subsection}}}

\begin{table}[h!]
\subsection{\textbf{Size of train, val and test sets.}}
\small
\setlength{\tabcolsep}{4.9pt}
\renewcommand{\arraystretch}{1.6}
\begin{tabular}{p{1.9cm}ccc|cr}
 & & \multicolumn{2}{c}{\textbf{Patches}} & \multicolumn{2}{c}{\textbf{Polygons}} \\ \cline{3-4} \cline{5-6}
\textbf{Set}  & \textbf{Area (km²)}      & \textbf{N}   & \textbf{\%} & \textbf{N} & \textbf{\%} \\ \hline
Train & 172.78 & 69111 & 51.0 & 330 & 73.5 \\
Val & 33.81 & 13523 & 10.0 & 58 & 12.9 \\
Test & 132.34 & 52935 & 39.0 & 61 & 13.6 \\ \hline
\label{appendix:trainvaltest}
\end{tabular}
\end{table}

\begin{table}[h!]
\subsection{\textbf{Support of semantic classes in train, val and test sets.}}
\small
\setlength{\tabcolsep}{4.9pt}
\renewcommand{\arraystretch}{1.6}
\begin{tabular}{p{1.9cm}ccc|cc|cc}
 & & \multicolumn{2}{c}{\textbf{Train}} & \multicolumn{2}{c}{\textbf{Val}} & \multicolumn{2}{c}{\textbf{Test}} \\ \cline{3-4} \cline{5-6} \cline{7-8}
\textbf{Class}              & \textbf{ID}       & \textbf{Patches}   & \textbf{Polygons} & \textbf{Patches} & \textbf{Polygons}   & \textbf{Patches} & \textbf{Polygons} \\ \hline
Deciduous oak & 0 & 15840 & 63 & 4374 & 14 & 27841 & 14\\ 
Evergreen oak & 1 & 11609 & 26 & 372 & 4 & 10380 & 6\\ 
Beech & 2 & 7008 & 21 & 1626 & 4 & 4036 & 4\\ 
Chestnut & 3 & 3337 & 16 & 147 & 2 & 200 & 3\\ 
Black locust & 4 & 1663 & 83 & 323 & 12 & 317 & 12\\ 
Maritime pine & 5 & 4568 & 20 & 960 & 3 & 2040 & 4\\ 
Scotch pine & 6 & 11330 & 34 & 2429 & 7 & 4506 & 5\\ 
Black pine & 7 & 4356 & 16 & 942 & 3 & 1928 & 3\\ 
Aleppo pine & 8 & 4028 & 15 & 233 & 2 & 438 & 2\\ 
Fir & 9 & 96 & 2 & 722 & 1 & 22 & 1\\ 
Spruce & 10 & 2579 & 16 & 627 & 3 & 868 & 4\\ 
Larch & 11 & 2536 & 7 & 503 & 1 & 255 & 1\\ 
Douglas & 12 & 161 & 11 & 265 & 2 & 104 & 2\\ \hline
\label{appendix:supports}
\end{tabular}
\end{table}

\begin{table}[h!]
\subsection{\textbf{Support of tree species in train, val and test sets.}}
\small
\setlength{\tabcolsep}{4.9pt}
\renewcommand{\arraystretch}{1.6}
\begin{tabular}{p{1.9cm}cp{1.9cm}cc|cc|cc}
 & & & \multicolumn{2}{c}{\textbf{Train}} & \multicolumn{2}{c}{\textbf{Val}} & \multicolumn{2}{c}{\textbf{Test}} \\ \cline{4-5} \cline{6-7} \cline{8-9}
\textbf{Class}              & \textbf{ID} & \textbf{Tree species}       & \textbf{Patches}   & \textbf{Polygons} & \textbf{Patches} & \textbf{Polygons}   & \textbf{Patches} & \textbf{Polygons} \\ \hline
\multirow{4}{*}{Deciduous oak} & \multirow{4}{*}{0} & Quercus robur & 144 & 4 & 1 & 1 & 302 & 1\\ 
 &  & Quercus pubescens & 12937 & 51 & 4084 & 11 & 27496 & 11\\ 
 &  & Quercus petraea & 2749 & 6 & 279 & 1 & 43 & 2\\ 
 &  & Quercus rubra & 10 & 2 & 10 & 1 & 0 & 0\\ \cline{1-3}
Evergreen oak & 1 & Quercus ilex & 11609 & 26 & 372 & 4 & 10380 & 6\\ \cline{1-3}
Beech & 2 & Fagus sylvatica & 7008 & 21 & 1626 & 4 & 4036 & 4\\ \cline{1-3}
Chestnut & 3 & Castanea sativa & 3337 & 16 & 147 & 2 & 200 & 3\\ \cline{1-3}
Black locust & 4 & Robinia pseudoacacia & 1663 & 83 & 323 & 12 & 317 & 12\\ \cline{1-3}
Maritime pine & 5 & Pinus pinaster & 4568 & 20 & 960 & 3 & 2040 & 4\\ \cline{1-3}
Scotch pine & 6 & Pinus sylvestris & 11330 & 34 & 2429 & 7 & 4506 & 5\\ \cline{1-3}
\multirow{2}{*}{Black pine} & \multirow{2}{*}{7} & Pinus nigra laricio & 1824 & 7 & 916 & 2 & 1288 & 2\\ 
 &  & Pinus nigra & 2532 & 9 & 26 & 1 & 640 & 1\\ \cline{1-3}
Aleppo pine & 8 & Pinus halepensis & 4028 & 15 & 233 & 2 & 438 & 2\\ \cline{1-3}
\multirow{2}{*}{Fir} & \multirow{2}{*}{9} & Abies nordmanniana & 29 & 1 & 0 & 0 & 0 & 0\\ 
 &  & Abies alba & 67 & 1 & 722 & 1 & 22 & 1\\ \cline{1-3}
Spruce & 10 & Picea abies & 2579 & 16 & 627 & 3 & 868 & 4\\ \cline{1-3}
Larch & 11 & Larix decidua & 2536 & 7 & 503 & 1 & 255 & 1\\ \cline{1-3}
Douglas & 12 & Pseudotsuga menziesii & 161 & 11 & 265 & 2 & 104 & 2\\ \hline
\end{tabular}
\label{tab:supports_species}
\end{table}

\end{document}